\documentclass[letterpaper, 10 pt, conference]{ieeeconf}  

\IEEEoverridecommandlockouts                              

\title{\LARGE \bf
CCO-VOXEL: Chance Constrained Optimization over Uncertain Voxel-Grid Representation for Safe Trajectory Planning } 

\usepackage{amsmath}
\usepackage{mathtools}
\usepackage{booktabs}
\usepackage{float}
\usepackage{xspace}
\usepackage{cite}
\usepackage{graphicx}
\graphicspath{{./figures/}}
\usepackage{caption}
\usepackage{subcaption}
\usepackage[inline]{asymptote}
\usepackage{tikz}
\usepackage{comment}
\usepackage{xcolor}
\usepackage{hyperref}
\let\labelindent\relax

\usepackage{enumitem}
\usepackage{amsthm}

\usepackage{nomencl}
\makenomenclature

\usepackage[noend]{algpseudocode}
\usepackage[linesnumbered,ruled]{algorithm2e}

\hypersetup{
    colorlinks=true,
    urlcolor=blue
    }
\usepackage{xcolor}
\PassOptionsToPackage{hyphens}{url}\usepackage{hyperref}

\usetikzlibrary{bayesnet}
\usetikzlibrary{arrows, positioning}

\usepackage{array}
\newcommand{\PreserveBackslash}[1]{\let\temp=\\#1\let\\=\temp}

\newcolumntype{C}[1]{>{\PreserveBackslash\centering}p{#1}}
\newcolumntype{L}[1]{>{\PreserveBackslash\raggedright}p{#1}}

\author{Sudarshan S Harithas$^{1}$, Rishabh Dev Yadav$^{1}$, Deepak Singh$^{1}$, Arun Kumar Singh$^{2}$, K Madhava Krishna$^{1}$

\thanks{$^{1}$are with RRC, IIIT Hyderabad, India
        {\tt mkrishna@iiit.ac.in, \{sudarshan.s, rishabhdev.yadav\}@research.iiit.ac.in}, {\tt say2deepaksingh@gmail.com}}%
\thanks{$^{2} $is with University of Tartu, Estonia 
        {\tt\footnotesize aks1812@gmail.com}}%

\thanks{The authors thank the anonymous reviewers for their helpful comments and \textit{Ministry of Electronics and Information Technology} (MeitY), Government of India, for their generous financial support. This work was supported in part by the European Social Fund via IT Academy program in Estonia, and grant PSG753 from Estonian Research Council. }

\thanks{$^\dagger$Project page: \href{https://github.com/sudarshan-s-harithas/CCO-VOXEL}{https://github.com/sudarshan-s-harithas/CCO-VOXEL}} 
}

\begin{document}

\makeatletter
\let\@oldmaketitle\@maketitle
\renewcommand{\@maketitle}{\@oldmaketitle
\centering
\vspace{-1mm}

\includegraphics[width=\textwidth]{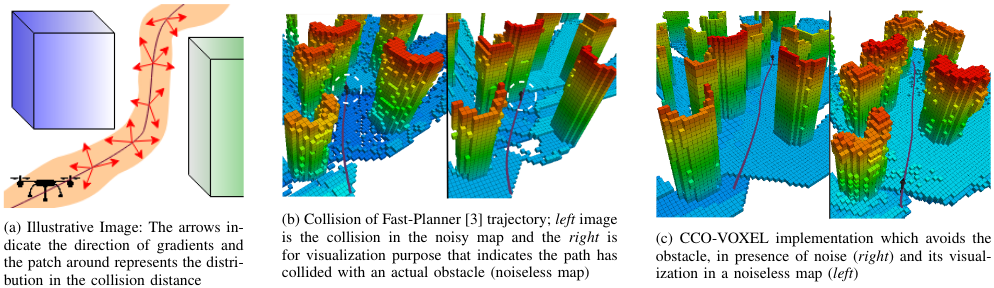}
\vspace{-5.5mm}
\captionof{figure}{\small{ \textbf{Overview}: Accurate gradient and distance to closest obstacles cannot be extracted from uncertain voxel grid representation built from noisy sensors. Fig.(a) illustrates this effect, wherein the gradient at a given point can be in any one of the directions shown by arrows. Similarly, the distribution of distance is shown as a confidence interval band. The noisy gradients and distance estimation tend to guide the trajectory towards the obstacle leading to a collision as shown in (b) (the collision is highlighted by a white circle) . We propose CCO-VOXEL, a method that accounts for the uncertainty in the distance measurements and obtains an optimal solution through a gradient free  \textit{Cross Entropy Minimization} technique resulting in a safe and optimal trajectories as shown in (c). }
}
\label{teaser}
\vspace{-5mm}
}

\newcommand\copyrighttext{%
  \footnotesize \textcopyright 2022 IEEE. Personal use of this material is permitted.
  Permission from IEEE must be obtained for all other uses, in any current or future
  media, including reprinting/republishing this material for advertising or promotional
  purposes, creating new collective works, for resale or redistribution to servers or
  lists, or reuse of any copyrighted component of this work in other works}

\newcommand\copyrightnotice{%
\begin{tikzpicture}[remember picture,overlay]
\node[anchor=south,yshift=10pt] at (current page.south) {\fbox{\parbox{\dimexpr\textwidth-\fboxsep-\fboxrule\relax}{\copyrighttext}}};
\end{tikzpicture}%
}

\makeatother

\maketitle
\hypersetup{urlcolor=black}

\copyrightnotice

\thispagestyle{empty}

\pagestyle{empty}

\begin{abstract}

We present CCO-VOXEL: the very first chance-constrained optimization (CCO) algorithm that can compute trajectory plans with probabilistic safety guarantees in real-time directly on the voxel-grid representation of the world. CCO-VOXEL maps the distribution over the distance to the closest obstacle to a distribution over collision-constraint violation and computes an optimal trajectory that minimizes the violation probability. Importantly, unlike existing works, we never assume the nature of the sensor uncertainty or the probability distribution of the resulting collision-constraint violations. We leverage the notion of Hilbert Space embedding of distributions and Maximum Mean Discrepancy (MMD) to compute a tractable surrogate for the original chance-constrained optimization problem and employ a combination of A* based graph-search and Cross-Entropy Method for obtaining its minimum. We show tangible performance gain in terms of collision avoidance and trajectory smoothness as a consequence of our probabilistic formulation vis a vis state-of-the-art planning methods that do not account for such non-parametric noise. Finally, we also show how a combination of low-dimensional feature embedding and pre-caching of Kernel Matrices of MMD allow us to achieve real-time performance in simulations as well as in implementations on on-board commodity hardware that controls the quadrotor flight.

\end{abstract}

\section{INTRODUCTION}
\vspace{-2mm}
Many aerial navigation systems compute trajectories over Octomap \cite{octomap}, or Euclidean Signed-distance field (ESDF) \cite{ESDF} derived from a three-dimensional voxel grid representation of the world. These approaches assume accurate sensor readings during the process of map creation. However, sensor measurements (e.g., point clouds) are noisy and often non-parametric in nature. These noisy measurements get translated to incorrect estimates of the distance to the closest obstacle and gradient field. Consequently, deterministic planners \cite{FastPlanner , RAPTOR, ewok , Topo} who rely on these two pieces of information being accurate are expected to lose their performance and safety guarantees under perception noise. For example, noisy gradients coupled with an incorrect estimate of the distance to the nearest obstacle can result in the UAV moving towards the obstacle and colliding. This paper proposes a chance-constrained optimization (CCO) over VOXEL grids as an antidote to this duress. CCO-VOXEL acknowledges that the distance to the closest occupied voxel is noisy and consequently takes this uncertainty into account while computing optimal trajectory plans. As a result, CCO-VOXEL avoids collisions where gradient-based deterministic planner fails (cf. \ref{teaser}). 


Previous methods leveraging CCO for handling sensor noise had to rely on the assumption that the underlying random variables that constitute the chance constraint function are parametric or Gaussian \cite{probabilistic_avoidance, IROS_15, bounding_volume, Mora_RAL2019}. However, as seen in Fig. (\ref{error_dist}) the distribution over the nearest voxel occupancy is non-parametric. Moreover, existing works on CCO such as \cite{Mora_RAL2019} assume that the obstacles have a pre-defined shape which is crucial for obtaining tractable reformulations of the chance constraints. In this paper, we relax both the assumptions of parametric uncertainty and the analytical collision model. Yet, we maintain real-time performance in both simulation and hardware experiments with on-board computation. The core innovations behind our approach are summarized below.

\noindent \textbf{Algorithmic Contribution:} For the first time, we present a CCO-based probabilistically safe trajectory planning that is agnostic to both the nature of the underlying uncertainty and the obstacle geometry. Inspired by our prior works \cite{Reactive_navigation_RKHS , CCO_RKHS}, we use the notion of distribution embedding in Reproducing Kernel Hilbert Space (RKHS) and Maximum Mean Discrepancy (MMD) to reformulate CCO as a distribution matching problem. However, we overcome one of the key limitations of \cite{Reactive_navigation_RKHS , CCO_RKHS}: estimating the so-called desired (or target) distribution for computing the distribution matching cost. We show that by embedding the distribution of constraint violation in RKHS instead of collision avoidance constraint themselves, as done in \cite{Reactive_navigation_RKHS , CCO_RKHS}, we can by-pass the need to estimate the desired distribution. Furthermore, distribution of constraint violation induces efficient structure in the algebraic form for MMD, allowing for pre-computation and caching of most computationally expensive parts. We also improve the computational performance of our approach by projecting the samples drawn from the distribution of constraint violations into some latent dimension before computing the MMD. We learn this projection through auto-encoder-based supervised learning. Finally, we minimize a combination of MMD and kinodynamic costs through the Cross-Entropy Method (CEM) to compute a smooth and probabilistically safe trajectory.

\noindent\textbf{State-of-the-art Performance:} We show tangible performance gain over deterministic gradient planners \cite{FastPlanner} and methods that handle uncertainty by growing the obstacle map by the co-variance of the distribution \cite{bounding1 ,bounding2}. The benefits are seen in the form of almost negligible collisions under the duress of varying levels of sensor noise even as the previous methods reports high collision rates.
\vspace{-2.0mm}

\section{Related Work}
\vspace{-2mm}
Trajectory planning for quadrotor systems has a long history in the areas of optimal trajectory generation \cite{gao2017quadrotor , probabilistic_avoidance, min_snap} and integrated perception and navigation \cite{polynomial_trajectory,Autonomous_exploration}. Gradient based planning over voxel representations have become popular in recent years \cite{FastPlanner, Topo} though they have not handled uncertainty in their formulations. Meanwhile, Chance Constrained Optimization with suitable surrogates too have been well studied, however their inadequacy in handling non-parametric uncertainty was detailed in \cite{Reactive_navigation_RKHS , CCO_RKHS}. Specifically there appears no known methods that handle non parametric uncertainty of discretized volumetric voxel grids other than the proposed. While \cite{CCO_distance_to_collision} provides for a chance constrained framework for nearest distance to obstacle it formulates over well defined obstacle shapes and not over voxel grids. Moreover, its computation of safe trajectories runs into minutes while the proposed method is in the order of tens of milliseconds. 

\renewcommand{\thefigure}{2}
\begin{figure}[]
	\centering
    \includegraphics[scale = 0.35]{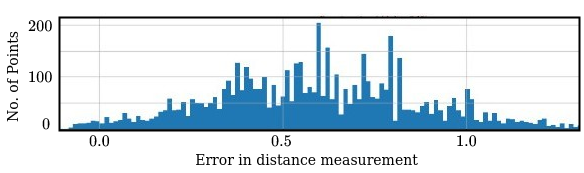} 
    \caption{\footnotesize The error (difference between the true collision distance and the measured value) distribution histogram that is obtained from a noisy voxel grid.  The OptiTrack Motion Capture system was used to determine the ground truth distance. Octomap with Euclidean Distance Transform (\href{https://github.com/OctoMap/octomap/tree/devel/dynamicEDT3D}{dynamicEDT3D}) is used for collision distance measurements. } \normalsize
    \label{error_dist}
\vspace{-7mm}
\end{figure}


\section{Problem Formulation} \label{Problem_formulation}

\noindent \textbf{Assumptions:} We assume that the robot has precise motion capabilities but noisy perception. As a result, the built ESDF has imprecise information about the gradient field and distance to the closest obstacle. The distribution over the distance has the form $d = d_{measured}+\epsilon_i$, where $\epsilon_i$ are samples drawn from some black-box distribution whose parametric form is not known.

\noindent \textbf{Trajectory Parametrization:} We parametrize quadrotor trajectory in the following manner
\vspace{-1mm}
\begin{equation}
\small
\begin{bmatrix}
x(t_1)\\
x(t_2)\\
\dots\\
x(t_n)
\end{bmatrix} = \textbf{P}\textbf{c}_{x}, \begin{bmatrix}
\dot{x}(t_1)\\
\dot{x}(t_2)\\
\dots\\
\dot{x}(t_n)
\end{bmatrix} = \dot{\textbf{P}}\textbf{c}_{x}, \begin{bmatrix}
\ddot{x}(t_1)\\
\ddot{x}(t_2)\\
\dots\\
\ddot{x}(t_n)
\end{bmatrix} = \ddot{\textbf{P}}\textbf{c}_{x}.
\label{param}
\vspace{-1mm}
\end{equation}

\noindent where, $\textbf{P}, \dot{\textbf{P}}, \ddot{\textbf{P}}$ are matrices formed with time-dependent basis functions (e.g polynomials) and $\textbf{c}_{x}$ are the coefficients associated with the basis functions. Similar expressions can be written for $y(t), z(t)$ as well in terms of coefficients $\textbf{c}_y, \textbf{c}_z$, respectively.  



\noindent \textbf{Safe Motion Planning:} Using the parametrization discussed above, we formulate our probabilistically safe motion planning as the following optimization problem, wherein $P(\cdot)$ represents probability.
\vspace{-4mm}
\begin{subequations}
   \begin{align}
       \min l(\textbf{c}_x, \textbf{c}_y, \textbf{c}_z) \label{cost} \\
        P( f(d) \leq  0 ) \geq \eta, \forall t \label{chance_const}
   \end{align}
\vspace{-2mm}
\end{subequations}
\vspace{-0.6cm}
\begin{align}
    f(d) = r_{safe}- d(\textbf{c}_x, \textbf{c}_y, \textbf{c}_z, t)
    \label{const_func}
\vspace{-4mm}
\end{align}


\noindent Our cost function (\ref{cost}) encourages smoothness in the trajectory by mapping the trajectory coefficients to penalties on higher-order motion derivatives such as jerk and penalising the violation of acceleration and velocity limit. The algebraic form is similar to that used in \cite{FastPlanner}. The function $d(\cdot )$ in (\ref{const_func}) maps the coefficients to distance to the closest obstacle at a given time $t$. Consequently, the inequality (\ref{chance_const}) ensures that the probability of the closest obstacle distance being less than safe value $r_{safe}$ is below some specified threshold $\eta$. Constraints of the form (\ref{chance_const}) are known as chance constraints \cite{chance_1}. Generally, chance-constrained optimization (CCO) are considered computationally intractable. Thus, existing works focuses on deriving tractable approximations of (\ref{chance_const}) under two predominant assumption. Firstly, the obstacles are assumed to have defined geometric shapes (e.g ellipsoid) and as a result it becomes possible to derive an analytical expression for $d(\textbf{c}_x, \textbf{c}_y, \textbf{c}_z, t)$. Second, the perception uncertainty are often assumed to be Gaussian. When both the assumptions are made simultaneously it even becomes possible to formulate a convex approximation of (\ref{chance_const}) \cite{Mora_RAL2019}, \cite{boyd_chance}.

Our work relaxes both the above mentioned assumptions and thus substantially expands the applicability of chance-constrained optimization based motion planning in real-world environment. We handle chance-constraints on collision avoidance by treating  $d(\textbf{c}_x, \textbf{c}_y, \textbf{c}_z, t)$ as a black-box model that can only be queried from a voxel-grid representation of the world. Secondly, we make no assumption on the underlying perception uncertainty.

\noindent \textbf{Reformulation from \cite{Reactive_navigation_RKHS}, \cite{CCO_RKHS} and Limitations:} Let $p_f$ represent the probability distribution of $f(d)$, i.e $p_f = P(f(d))$. Our prior work \cite{Reactive_navigation_RKHS}, \cite{CCO_RKHS} imagined CCO as the problem of choosing an appropriate $(\textbf{c}_x, \textbf{c}_y, \textbf{c}_z)$ such that the $p_f$ at any given time $t$ takes on an appropriate shape. This naturally gives rise to the notion of desired distribution $p_{f_{des}}$, i.e a distribution that $p_f$ should resemble as closely as possible (cf.  Fig.  \ref{dist_figure_ral}). The entire mass of the $p_{f_{des}}$ lies to the left of $f(d)=0$. As $p_f$ becomes similar to $p_{f_{des}}$, its mass starts getting shifted to the left and the probability of $P(f(d)\leq 0)$ goes up. This intuition led to the following reformulation of optimization
(\ref{cost})-(\ref{chance_const})
\vspace{-1.5mm}
\begin{align}
    \min l(\textbf{c}_x, \textbf{c}_y, \textbf{c}_z)+wl_{dist}(p_f, p_{f_{des}}),
    \label{dist_matching_ral}
\vspace{-3.0mm}
\end{align}
\noindent where, $l_{dist}$ is a cost that quantifies the similarity between two distributions. The constant $w$ controls the trade-off between minimizing the distribution similarity (or probability of collision avoidance) with the primary cost. We discuss suitable form for $l_{dist}$ in the next section.


The \textbf{main limitation} of \cite{Reactive_navigation_RKHS}, \cite{CCO_RKHS} is that it requires one to solve an optimization problem to estimate $p_{f_{des}}$. While tractable, in a reactive one-step setting, this estimation process becomes highly challenging for multi-step trajectory planning such as the one considered in this paper. In the next section, we introduce our main algorithmic results: a novel RKHS reformulation of the CCO (\ref{const_func})-(\ref{chance_const}) that does not require us to estimate $p_{f_{des}}$.


\renewcommand{\thefigure}{3}
  \begin{figure}
  \begin{subfigure}{0.49\columnwidth}
  \includegraphics[width=\textwidth]{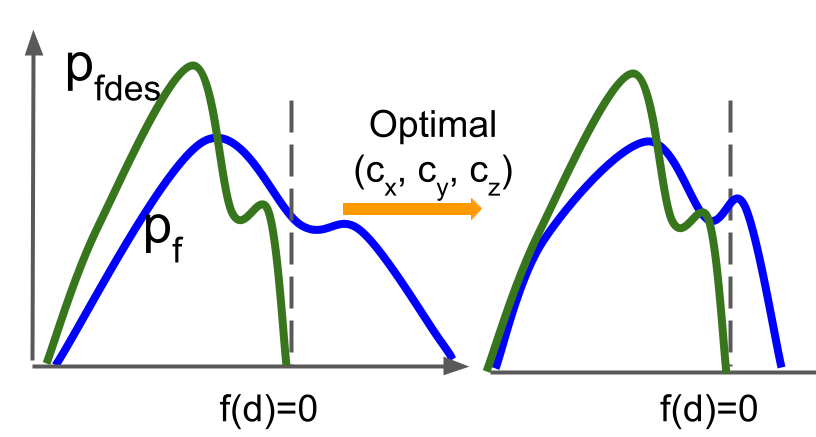}
  \caption{The setting from \cite{Reactive_navigation_RKHS} where the goal is to bring the distribution $p_{f}$ to the left of $f(d)= 0$. This is achieved by constructing a desired distribution $p_{f_{des}}$ and minimizing the distance of $p_f$ from it in RKHS. Importantly, estimation of  $p_{f_{des}}$ required solving an optimization problem  }
  \label{dist_figure_ral}
  \end{subfigure}
  \hfill
  \begin{subfigure}{0.49\columnwidth}
  \includegraphics[width=\textwidth]{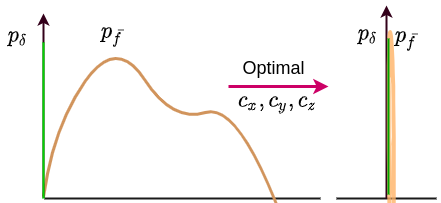}
  \caption{ We work with the distribution of the collision constraint violation rather than the constraint themselves. As a result, the desired distribution in our case is simply the Dirac Delta distribution $p_{\delta}$. The distribution  $p_{\overline{f}}$ approaches the desired Dirac function $p_{\delta}$ for an optimal $\textbf{c}_x, \textbf{c}_y, \textbf{c}_z$ } 
  \label{dirac_matching}
  \end{subfigure} 

  \caption{An intuitive understanding of the RKHS embedding of \cite{Reactive_navigation_RKHS} and our proposed method} 
 \vspace{-7mm}
  \end{figure}

\section{Main Algorithmic Results}

\subsection{Distribution over Constraint Violations}
\noindent Our main idea is to work with the distribution of violations of the collision avoidance constraints rather than the constraints themselves. More precisely, let $\overline{f}$ be the constraint violation function defined as
\vspace{-2mm}
\begin{align}
    \overline{f} = \max(0, f(d(\textbf{c}_x, \textbf{c}_y, \textbf{c}_z, t)))
    \label{const_viol_function}
\vspace{-3mm}
\end{align}
\noindent A distribution over $d$ can be mapped to $\overline{f}$ and we represent it as $p_{\overline{f}}$. Although, computing the analytical form for $p_{\overline{f}}$ is intractable, we can make the following remark regarding the best possible shape that it can take in the context of collision avoidance.

\newtheorem{remark}{Remark}
\begin{remark} \label{des_dirac}
The best possible shape of $p_{\overline{f}}$ is given by a Dirac Delta distribution $p_{\delta}$ . In other words, $p_{\delta}$ acts as the desired distribution for $p_{\overline{f}}$  
\end{remark}
\begin{remark}\label{dirac_sim}
As $p_{\overline{f}}$ becomes more and more similar to $p_{\delta}$, the probability of collision avoidance goes up.
\end{remark}

\noindent Remark \ref{des_dirac} and \ref{dirac_sim} elucidates our motivation behind shifting from the distribution of constraints, as done in \cite{Reactive_navigation_RKHS}, \cite{CCO_RKHS}, to distribution of constraint violations. The desired distribution in our case is always fixed and most importantly, exactly known. Remark \ref{dirac_sim} forms the basis of the following proposed reformulation of CCO.
\vspace{-2mm}
\begin{equation}
    \min_{\textbf{c}_x, \textbf{c}_y, \textbf{c}_z} l(\textbf{c}_x, \textbf{c}_y, \textbf{c}_z)+w l_{dist}(p_{\overline{f}}, p_{\delta})
    \label{dist_matching_proposed}
\vspace{-2mm}
\end{equation}

\noindent Note that $p_{\overline{f}}$ is in fact a function of $d$ which in turn is a function of the trajectory coefficients $(\textbf{c}_x, \textbf{c}_y, \textbf{c}_z)$

\subsection{Maximum Mean Discrepancy and Sample Approximation} 
\noindent A common method to quantify similarity between two distributions is through Kullback Liebler Divergence (KLD). However, it requires us to know the analytical form for the distributions under consideration and thus is not suitable to quantify the similarity between $p_{\overline{f}}$ and $p_{\delta}$. One possible workaround is provided by the notion of Maximum Mean Discrepancy (MMD) that quantifies the similarity of two distributions in Reproducing Kernel Hilbert Space (RKHS). Importantly, both the embedding into RKHS and MMD can be obtained by just sample level information of distributions. To this end, let $\mu_{p_{\overline{f}}}$ and $\mu_{p_{\delta}}$ represent the RKHS embedding of $p_{\overline{f}}$ and $p_{\delta}$ computed in the following manner.
\noindent\begin{minipage}{.5\linewidth}
\vspace{-3mm}
\begin{equation}
 \small
\mu_{p_{\overline{f} }} = \sum_{i=0}^{N} \alpha_{i}k( \overline{f}(d_{i}), \cdot) \label{kernel_mean_fover}
\end{equation}
\end{minipage}%
\begin{minipage}{.5\linewidth}
\vspace{-3mm}
\begin{equation}
\mu_{p_{\delta}} = \sum_{i=0}^{N} \beta_{i}k(0,  \cdot) \label{kernel_mean_dirac}
\end{equation}
\end{minipage}


\noindent where, $d_i$ refers to the $i^{th}$ sample drawn from the distribution of $d$ for a given $(\textbf{c}_x, \textbf{c}_y, \textbf{c}_z, t)$. Please note that (\ref{kernel_mean_dirac}) follows from the fact the samples from a Dirac Delta distribution are all zeroes. Using (\ref{kernel_mean_fover}) and (\ref{kernel_mean_dirac}), we can formulate $l_{dist}$ in the following manner.
\vspace{-5mm}
\begin{equation}
\small    l_{dist} = \left\Vert \mu_{p_{\overline{f}}} - \mu_{p_{\delta}} \right \Vert_{2}^{2}
    \label{mmd_ldist}
\vspace{-1mm}
\end{equation}


\noindent It is worth reiterating that MMD (\ref{mmd_ldist}) is a function of $(\textbf{c}_x, \textbf{c}_y, \textbf{c}_z, t)$. Each possible choice leads to a different distribution of $d$ and consequently MMD. Thus the goal is to come up with the right trajectory coefficients which minimizes both MMD and the primary cost function. We achieved this through Cross-Entropy Method \cite{CEM1} \cite{iCEM} and is discussed in Sec. \ref{CEM_content}.
 \vspace{-2mm}

\subsection{Pre-Computation and Improving Sample Complexity} \label{MMD_formulation}

\noindent \textbf{Computing MMD through Kernel Matrices:} Let ${d_{0} , d_{1}, d_{2} , \dots , d_{n}}$ be the samples drawn from the distribution of the closest obstacle at any given time $t$ for a given $\textbf{c}_x, \textbf{c}_y, \textbf{c}_z$. The distance samples can then be mapped to samples from $p_{\overline{f}}$ through (\ref{const_viol_function}). In this part, we derive the matrix representation of the MMD (\ref{kernel_mean_dirac}) to show how various parts of it can be pre-computed and cached. To this end, we expand (\ref{mmd_ldist}) in the following manner.
\vspace{-1.5mm}
\begin{align} 
\small
    \lVert \mu_{p_{\overline{f}}} -  \mu_{p_{\delta}} \lVert ^{2} = & \langle {\mu_{p_{\overline{f}}}(d)} , {\mu_{p_{\overline{f}}}(d)} \rangle - 2 \langle {\mu_{p_{\overline{f}} }}, {\mu_{p_{\delta}}} \rangle + \nonumber \\ & \langle {\mu_{p_{\delta}}} , {\mu_{p_{\delta}}} \rangle
\label{MMD_expand}
\vspace{-2mm}
\end{align}



Substituting kernel mean functions (\ref{kernel_mean_fover}) and (\ref{kernel_mean_dirac}) in (\ref{MMD_expand})

\vspace{-2mm}
\begin{subequations}
\small
\begin{gather}
  \langle  \mu_{p_{\overline{f}}}(d)  ,  \mu_{p_{\overline{f}}} \rangle = \langle  \sum_{i=0}^{N}  { \alpha_{i}k( \overline{f}(d_{i}), \cdot } , \sum_{j=0}^{N} {\alpha_{j}k( \overline{f}(d_{j}), \cdot } \rangle \\
\langle  \mu_{p_{\overline{f}}}(d) , \mu_{p_{\delta}}  \rangle = \langle \sum_{i=0}^{N} { \alpha_{i}k( \overline{f}(d_{i}), \cdot }  , \sum_{j=0}^{N} \beta_{j} {k(0, \cdot )} \rangle \\
  \langle  \mu_{p_{\delta}} ,\mu_{p_{\delta}} \rangle = \langle \sum_{i=0}^{N}  {\beta_{i}k(0, \cdot )}  , \sum_{j=0}^{N} {\beta_{j}k(0, \cdot )} \rangle
\end{gather}
\label{MMD_main} 
\vspace{-2mm}
\end{subequations}
 
 Using the kernel trick Eq. (\ref{MMD_main}) can be reduced to a matrix form and can be expressed as 
 \vspace{-1mm}
  \begin{equation} \label{MMD_matrix}
  \small
   \lVert \mu_{p_{\overline{f}}} -  \mu_{p_{\delta}} \lVert ^{2}  = \mathbf{C_{\alpha}K_{\overline{f}\overline{f}}C_{\alpha}^{T} - 2 C_{\alpha}K_{ \overline{f}0}C_{\beta}^{T} + C_{\beta}K_{00}C_{\beta}^{T} } 
 \vspace{-1mm}
\end{equation}
where $C_{\alpha}$ and $C_{\beta}$ are the weight vectors given by
\vspace{-1mm}
\begin{gather*}
 C_{\alpha} = \begin{bmatrix}
\alpha_{0} , \alpha_{1} , ... , \alpha_{n}
\end{bmatrix}; 
C_{\beta} = \begin{bmatrix}
\beta_{0} , \beta_{1} , ... , \beta_{n}
\end{bmatrix} 
\vspace{-2mm}
\end{gather*}
and $K_{\overline{f}\overline{f}}$, $K_{ \overline{f}\delta}$ and  $K_{\delta\delta}$ are matrices defined as
\vspace{-1mm}
\[
\footnotesize
\mathbf{K_{\overline{f} \overline{f}}} = \begin{bmatrix}
 k( \overline{f}(d_{0})  , \overline{f}(d_{0}) ) & \hdots & k( \overline{f}(d_{0}) ,\overline{f}(d_{n}) ) \\
 k(\overline{f}(d_{1}) , \overline{f}(d_{0}) ) & \hdots & k( \overline{f}(d_{1}) , \overline{f}(d_{n}) )\\
 \vdots                  & \ddots    & \vdots & \\ 
 k( \overline{f}(d_{n})  , \overline{f}(d_{0}) )  & \hdots & k(\overline{f}(d_{n}) , \overline{f}(d_{n}) ) \\
 \end{bmatrix}  \]
\[ 
\footnotesize
\mathbf{K_{f\delta}} = \begin{bmatrix}
k( \overline{f}(d_{0}) , 0) & \hdots & k( \overline{f}(d_{0}) , 0) \\
k( \overline{f}(d_{1}) , 0)  &  \hdots & k(\overline{f}(d_{1}) ,0) \\
\vdots                  & \ddots    & \vdots & \\ 
k( \overline{f}(d_{n}) ,0) &  \hdots & k( \overline{f}(d_{n}) , 0)
\vspace{-2mm}
\end{bmatrix} \]

\[ \mathbf{K_{\delta\delta}} = \mathbf{1}_{n\times n} \vspace{-2mm}\]
We use Radial Basis Functions (RBF) during implementation. 
The use of the Dirac-Delta function as the desired function leverages us with two key properties which would further decrease the computation time. 
\begin{enumerate}
    \item Pre computing matrices: The matrix $K_{\delta \delta}$ is essentially a set of ones and it does not need explicit computation at run time. Furthermore, if a polynomial kernel is used the matrix $K_{\Bar{f} \delta}$ would also reduce to set of ones and only one matrix needs computation at run time. 
    \item Symmetric matrix: It can be observed the kernel matrices are square and symmetric, this implies that we can compute values only for one triangle of the matrix and copy the values into the other triangle. 
\end{enumerate}
The use of these techniques in addition to feature space embedding, and vectorized matrix-algebra enables the computation of MMD in real-time.

\noindent \textbf{Feature space embedding of distance measurements:} We need a total of $n$ samples of $\overline{f}$ to compute its embedding in RKHS and consequently MMD. A large $n$ $(\approx 100)$ gives a good estimate of MMD but at the same time increases the computational burden (the size of Kernel matrix in (\ref{MMD_matrix}) increases). In this part, we use an auto-encoder to improve the sample and computation complexity. More concretely, we stack the $n$ samples of  $\overline{f}$ at a given point on the trajectory and project it into a low dimensional latent space. Our auto-encoder based embedding can be seen as the more sophisticated version of reduced-set method used in \cite{Reactive_navigation_RKHS}.

We learn the low-dimensional mapping through supervised learning. Formally, we seek to determine an encoder \textbf{$W_{1}$} and decoder \textbf{$W_{2}$} pair such that they perform an identity operation.  The training is performed by optimizing the loss function given in  (\ref{autoencoder:a}) using \textit{Stochastic Gradient Descent}.The training data consists of $\mathnormal{n}$ sets of $\overline{f}$, each with $\mathnormal{m}$ samples, stacked to form the data matrix $\textbf{D}$ of dimension $\boldsymbol{R^{n \times m }}$.
\vspace{-3mm}
  \label{autoencoder}
    \begin{align}
      \min_{ \textbf{W}_1 ,\textbf{W}_2} \left \Vert \textbf{D}\textbf{W}_1\textbf{W}_2-\textbf{D} \right \Vert^{2} \label{autoencoder:a} 
\vspace{-5mm}
\end{align}
\noindent The final latent space dimension where MMD would be computed is given by $\textbf{D}\textbf{W}_{1}$


\vspace{-2mm}

\subsection{Trajectory Optimization}
\noindent We minimize (\ref{dist_matching_proposed}) in  two steps. At the first step, we use graph-search technique to find a trajectory that approximately minimizes (\ref{dist_matching_proposed}). This resulting trajectory is used to initialize the Cross Entropy Method (CEM) optimizer for computing an optimal solution of (\ref{dist_matching_proposed}).

\noindent \textbf{Initial Trajectory Search:} 
Our initial trajectory search builds on Kinodynamic A* proposed in \cite{FastPlanner} but with a crucial difference that we use MMD as a part of the \textit{Edge Cost} that connects two nodes of the graph. The typical process is as follows. Several motion primitives are generated by sampling different control inputs $\textbf{u}_t$ (acceleration). These primitives are then assigned an edge-cost (\ref{edge_cost}), wherein $\Delta M$ (\ref{delta_mmd}) is the difference between MMD at the current, and the next node and $\tau$ is the time for which the control input is applied. The constant $\rho$ trades-off control costs with time. Apart from the edge cost, there is also a heuristic cost that measures the goal-reaching aspect of each primitive. The primitive with the best edge and heuristic cost combination is chosen for expansion in the next iteration.

In contrast to our approach, \cite{FastPlanner} uses the distance to the closest obstacle (or simply occupancy in an inflated map) to filter out unsafe primitives while the edge-cost only has the first two terms from (\ref{edge_cost}). As mentioned earlier, if this distance information is noisy, the A* process itself can lead to unsafe trajectories (see Table \ref{front_end_comparision}). 
\vspace{-2mm}


\noindent\begin{minipage}{.5\linewidth}
\begin{equation}
 \small
 e_{c} = ( \lVert u_{t} \lVert^{2} + \rho )\tau + \Delta M  \label{edge_cost} 
\end{equation}
\end{minipage}%
\begin{minipage}{.5\linewidth}
\begin{equation}
 \Delta M  = M_{i+1} - M_{i} \label{delta_mmd}
\end{equation}
\end{minipage}

\vspace{1mm}

\noindent \textbf{CEM Based Refinement:} \label{CEM_content}
The trajectory produced by the graph search can be sub-optimal and might have a higher arc length. We use Cross Entropy Minimization (CEM) to further refine the trajectory and improve its smoothness while maintaining clearance from the obstacle. The various steps of CEM are summarized in Algorithm \ref{ICEM_algo}. The input to the CEM is the sub-optimal trajectory obtained from graph-search. The initial trajectory is converted into an initial estimate for trajectory coefficients $(\textbf{c}_x, \textbf{c}_y, \textbf{c}_z)$ through simple curve-fitting (line \ref{FitPolynomial}). The first step in Algorithm \ref{ICEM_algo} is drawing $N$ samples of trajectory coefficients, leading to $N$ possible trajectories (line \ref{DrawSamples}). At each point of every sampled trajectory, we computed the cost function (\ref{dist_matching_proposed}) (line \ref{DetermineCost}). Note that as shown section \ref{MMD_formulation}, the MMD evaluation itself requires drawing $n$ samples from the collision distance distribution at each point of all the sampled $N$ trajectories. We select the top $q$ trajectory coefficient samples (line \ref{SelectElites} ) that led to lowest cost and use them to compute the mean and variance for the next iteration of CEM (line \ref{FitGaussian}). The output of the CEM is the coefficients of the optimal trajectory.



In a standard CEM implementation after every iteration of the inner loop (line \ref{DrawSamples} to \ref{SelectElites}) all the elite samples are discarded. This decreases the efficiency and increases the convergence time. Authors in \cite{iCEM} propose \textit{CEM with memory} where a small subset of the elite set is stored and added to the pool of samples for the next iteration. Our CEM uses this insight in line \ref{AddSamples} of Algorithm \ref{ICEM_algo}.

\vspace{-2mm}
\begin{algorithm}
\caption{CEM based Trajectory Refinement}
\label{ICEM_algo}
\SetAlgoLined
\textbf{Input:} Sub optimal trajectory ($x(t), y(t), z(t)$) from initial path search and trajectory execution time $t_{exe}$\\
$N_{CEM}$ = Number of CEM interations\\
$\textbf{c}_{x}, \textbf{c}_y, \textbf{c}_z$  = Trajectory coefficients\\
$\textbf{c}_{x}, \textbf{c}_y, \textbf{c}_z \gets \textbf{FitPolynomial}($x(t), y(t), z(t)$) $\label{FitPolynomial} \\
$\boldsymbol{\mu}_{0} = (\textbf{c}_{x}, \textbf{c}_y, \textbf{c}_z)$\\
\For{$i=0, i \leq N_{CEM}, i++$}
{
Initialize $CostList$ = []
\\$Pool = \text{N samples from }\mathcal{N}( \mu_{i} , diag( \sigma_{i} ^{2} ))$ \label{DrawSamples}\\
    \For{$j=0, j \leq \text{size}(Pool), j++$}
    {
    $cost \gets \textbf{DetermineCost}(S_j)$ \label{DetermineCost}\\ 
    append ${cost}$ to $CostList$
    }
$EliteSet  \gets \textbf{SelectElites}(CostList, Pool)$ \label{SelectElites}\\
$Pool \gets \textbf{UpdateSamples}(EliteSet)$ \label{AddSamples}  \\
$(\mu_{i+1}, \sigma_{i})  \gets  \text{fit Gaussian distribution }EliteSet $ \label{FitGaussian}\\
$\textbf{c}_{x}, \textbf{c}_y, \textbf{c}_z = \boldsymbol{\mu}_{i}$
}
$\text{Optimized Trajectory} \gets \mathbf{P}\mathbf{c_{x}}, \textbf{P}\textbf{c}_y, \textbf{P}\textbf{c}_z$\\
\end{algorithm}
 \vspace{-5mm}

\section{Experiments and Validation}

In this section, we demonstrate the advantages of CCO-VOXEL over deterministic methods and other conservative approaches to handle uncertainty through qualitative and quantitative comparisons. We further show that with the convergence of CEM to an optimal solution the collision constraint violation distribution will also converge to the dirac delta function. Ablations confirm the real-time competency of CCO-VOXEL, while the actual experimental runs indicate its robustness and efficacy for on-board implementations. 

\noindent \textbf{Benchmark Comparisons}: We benchmark our CCO-VOXEL against fast planner \cite{FastPlanner} and bounding volume \cite{ bounding1 , bounding2}  algorithms. The comparison is done on two highly cluttered simulation environments named \textit{Box Cylinder world}  and \textit{ Wall Grid World} (cf. Fig. \ref{Gazebo_world}) of dimensions $ 30 \times 30 \times 7$ and $ 7 \times 30 \times 7$ respectively. Over $200$ trials were conducted and the results presented in Table \ref{benchmark} are the cumulative statistics of these trials in both the testing environments. Maximum velocity and acceleration were set to $2 m/s$ and $3 m/s^{2}$ respectively. We sample $50$ random trajectories at every iteration of CEM. We implemented CCO-VOXEL in C++11, and used the PX4 SITL (Software in the Loop) for simulations. During simulation, a Gaussian noise of zero mean and $0.2m$ variance is added into the point cloud for testing purposes. \textbf{Note} that although the point cloud noise is Gaussian, its mapping to the distribution of the closest obstacle will be non-parametric. 

The Fast-Planner \cite{FastPlanner} is a state-of-the-art algorithm known to generate optimal and safe trajectories. However, in a cluttered and noisy environment the distance to obstacle measurements are no longer reliable. Furthermore, the gradients are noisy and can push the trajectory towards the obstacle leading to a collision (cf. Fig. \ref{teaser}{\color{red}(a)} and \ref{teaser}{\color{red}(b)}) and a drop in the success rate and increase in the smoothness cost of the trajectory. On other hand, CCO-VOXEL is gradient free, and the MMD formulation takes into account the noise uncertainty in the collision distance (cf. Fig. \ref{teaser}{\color{red}(c)}). As a result, it achieves a higher success rate and a lower smoothness cost.

A possible approach to account for uncertainty is to construct a bounding volume around the obstacle, which is obtained by increasing the robot's footprint by the size of the covariance of the distance distribution. This technique has been used in several approaches such as \cite{bounding1}. Due to the increase in a safety margin, the bounding volume approaches have a higher accuracy in comparison to the fast planner \cite{FastPlanner}, but they tend to generate conservative trajectories. Furthermore, this method prevents the planner from navigating in narrow spaces where it is difficult to maintain an appropriate safe margin after the covariance-based inflation.

  
As shown in Table \ref{benchmark} the proposed method out performs both the benchmarks by a fair margin. The results of our experiments demonstrate a \textbf{33.17\%} and \textbf{44.94\%} decrease in the overall trajectory smoothness cost in comparison to Fast-Planner and bounding volume based approaches respectively. Furthermore, the success rate of CCO-VOXEL is approximately \textbf{3} times higher in comparison to the benchmarked approaches. But our method has a slightly higher computation time, due to the relatively high computation requirement of the evaluating the MMD cost in CEM. 




\noindent \textbf{Real World Autonomous Flight} We have conducted a fully autonomous flight test in unknown and cluttered environment. The demonstration results can be found \href{https://github.com/sudarshan-s-harithas/CCO-VOXEL/tree/main/Demo}{here}\footnote{\href{https://github.com/sudarshan-s-harithas/CCO-VOXEL/tree/main/Demo}{https://github.com/sudarshan-s-harithas/CCO-VOXEL/tree/main/Demo}}. We used a customized self developed quadrotor platform, which is equipped with an Intel Realsense D415 depth camera \cite{realsensecamera}, a Pixhawk flight controller is used along with the PX4 firmware, all the computation  modules which include mapping, state estimation and motion planning is conducted online on a Intel NUC 10 board \cite{nuc} with Core i7-10710U Processor,1.1 GHz – 4.7 GHz Turbo, 6 core, the system is equipped with a 16GB RAM and 1TB SSD. 

\begin{table}[]
\vspace{+2mm}
\caption{Benchmark Comparison}
\centering  
\begin{tabular}{l c  rr}  
\hline\hline                       
\\ [-2ex]
Method  
&  Computation & Smoothness & Success \\ 
& Time(s) & ($m^2/s^5$) & \% \\ [0.5ex]
\hline
\\ [-2ex]

&mean: $0.0853$ & $\mathbf{12.45}$ & $\mathbf{100}$  \\ [-1ex]
\raisebox{1.5ex}{Proposed} & std: $0.014$ &   &  \\[1.0ex]

&mean: $0.012$ & $18.63$ & $32.37$  \\ [-1ex]
\raisebox{1.5ex}{FastPlanner} & std: $0.026$ &   &  \\[1.0ex]

&mean: $0.007$ & $22.54$ & $36.76$  \\ [-1ex]
\raisebox{1.5ex}{Bounding Volumes} & std: $0.001$ &   &  \\[0.5ex]

\hline                          
\end{tabular}
\label{benchmark}
\end{table}
\vspace{-6mm}




\renewcommand{\thefigure}{4}
  \begin{figure}
  \vspace{-4mm}
  \begin{subfigure}{0.49\columnwidth}
  \includegraphics[width=\textwidth]{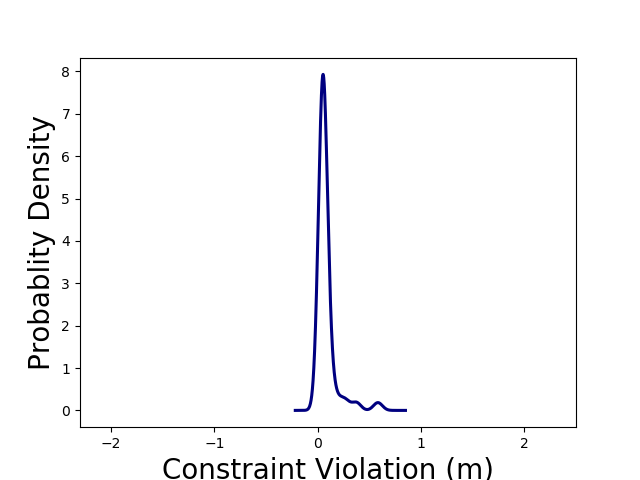}
  \caption{Iteration 0 : Initilization}
  \end{subfigure}
  \hfill
  \begin{subfigure}{0.49\columnwidth}
  \includegraphics[width=\textwidth]{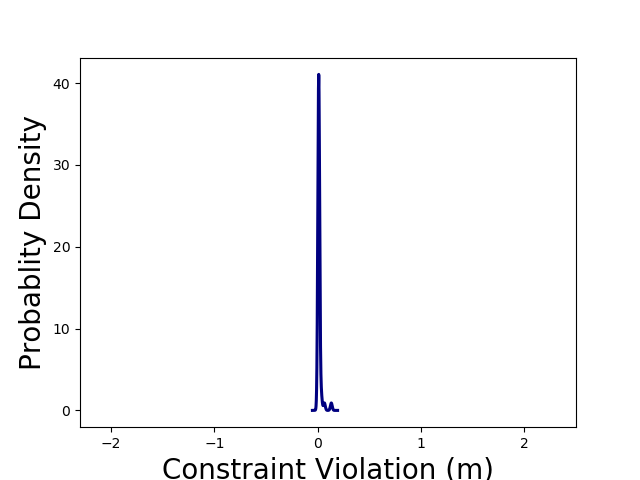}
  \caption{Iteration 5} 
  \end{subfigure} 
  %
  \caption{\footnotesize With every iteration of our CEM based trajectory optimization we observe that the distribution of the collision-constraint violation approaches the Dirac delta function and therefore the MMD cost is minimized. Note that since plotting is based on approximate kernel density estimation from finite samples, a tiny part of distribution appears to the left of 0 as well.} \label{MMD_convergence}
 \vspace{-4mm}
 \end{figure}







\vspace{4mm}
\subsection{Convergence analysis}
\vspace{-1mm}
\noindent \textbf{Cross Entropy Minimization}:  \label{CEM_convergence} In this section we demonstrate the ability of the CEM to compute the minimum of our unconstrained reformulation of CCO (\ref{dist_matching_proposed}). The CEM takes the output of the initial trajectory search and further refines it. The convergence of CEM is validated in Fig. \ref{CEM_optimization}. The figure presents the cost profile of the mean trajectory after every iteration; it can be seen that the optimal solution can be obtained in around four iterations with an average computation time less than $0.015s$. The convergence can be verified by observing the variance updates after every iteration. A monotonic decrease in variance implies that after every iteration it converges to low-cost region and thus it is more useful to sample trajectories close to mean trajectory.      

\noindent \textbf{Convergence of the Constraint Violation}: We demonstrate that the convergence of CEM (and minimization of MMD) implies a greater match between the distribution of collision constraint violation and the Dirac Delta function. This in turn leads to an increased probability of collision avoidance (recall Remark \ref{dirac_sim}). We observe from Fig. \ref{MMD_convergence} that five iterations were enough for the constraint violation distribution to converge to the Dirac distribution. This convergence results have been empirically verified over a large number of trials.

\renewcommand{\thefigure}{5}
\begin{figure}[!t]
\centering
\includegraphics[scale = 0.4]{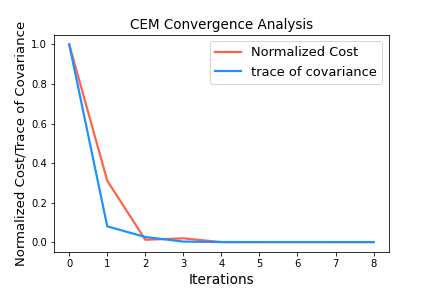}
	\caption{\footnotesize CEM Optimization: We observe that the normalized cost and the trace of the covariance has monotonically decreased and the trajectory has converged to an optimal solution }
		\label{CEM_optimization}
\vspace{-7mm}
\end{figure}

\vspace{-3mm}

\subsection{Ablation Study}
\vspace{-1mm}
\noindent To study the impact of each component of our approach, we conducted an ablation study between the possible methods.


\noindent \textbf{MMD Computation}: In Sec. \ref{MMD_formulation} we described the use of an auto-encoder architecture to decrease the sample complexity and consequently time required for an MMD computation. As presented in Table \ref{MMD_comparisions} we compare this approach against the baseline method where RKHS embedding  (Eq. \ref{MMD_matrix}) are directly computed for a large number of samples without any low dimensional projection. It can be observed that the use of an auto-encoder decreases the computation time by approximately \textbf{12} times. Furthermore, we observe from Fig.\ref{MMD_convergence} that the increase in computation time did not compromise the performance, we were still able to minimize the probability of constraint violations within a very few iterations of CEM.

\begin{table}[h] 
\vspace{-2mm}
\caption{Comparison between MMD computation methods }
\centering  
\begin{tabular}{l c  rr}  
\hline\hline                     
\\ [-2ex]
Method
&  Computation Time(s) \\ [0.5ex]
\hline
\\ [-1ex]
Autoencoder Embedding & $\mathbf{0.00012}$ \\[0.5ex]
Traditional Method & 0.00146\\ [0.5ex]
\hline                          
\end{tabular}
\label{MMD_comparisions}
 \vspace{-3mm}
\end{table}


\noindent \textbf{Initial Trajectory Searching Algorithm}: Here, we validate the effectiveness of using MMD as the edge-cost in A* based initial trajectory search. To this end, we compare it against the standard kinodynamic A* used in \cite{FastPlanner}. As shown in Table \ref{front_end_comparision}, while our inclusion of MMD as edge cost increases the computation time, it comes up with a massive \textbf{2.82} times improvement in success rate. It is worth recalling that our computation time is higher because MMD is computed over the several samples of closest distance to the obstacle drawn from a black-box distribution. 


\vspace{-2mm}


\begin{table}[h] \label{front_end_comparision}
\caption{Initial Trajectory Searching Comparison}
\centering  
\begin{tabular}{l c  rr}  
\hline\hline                     
\\ [-2ex]
Method
&  Computation Time(s) & Success \% \\ [0.5ex]
\hline
\\ [-1ex]
Kinodynamic A* with MMD & 0.0798 &  $\mathbf{100}$ \\[0.5ex]
Kinodynamic A* \cite{FastPlanner} & 0.017  &    35.4\\ [0.5ex]
\hline                          
\end{tabular}
\label{front_end_comparision}
 \vspace{-5mm}
\end{table}

\renewcommand{\thefigure}{6}
\begin{figure}
\vspace{+2mm}
\centering
\includegraphics[scale = 0.7]{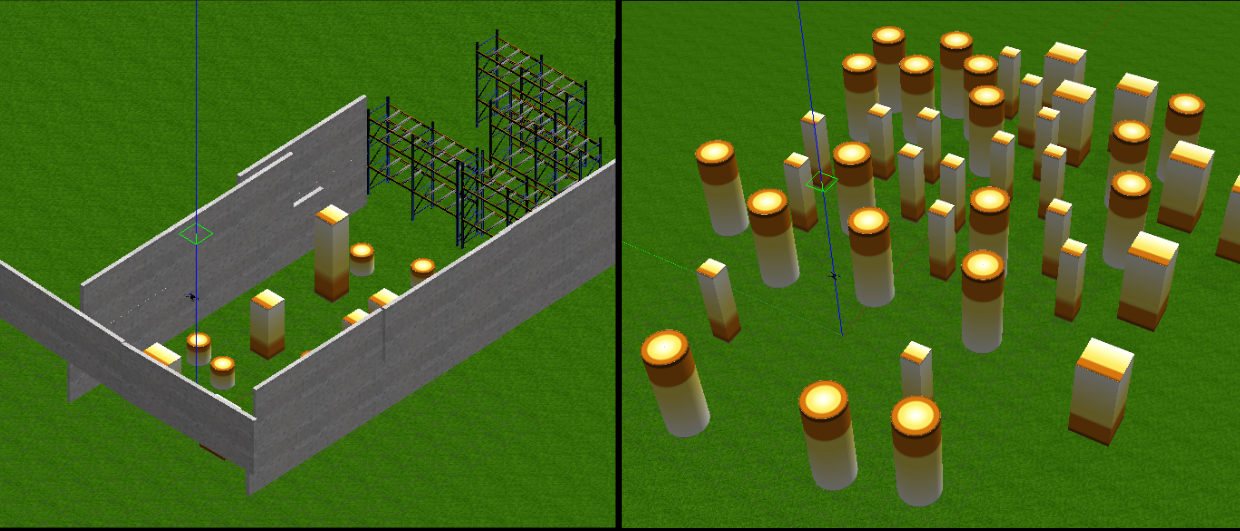}
	\caption{\footnotesize Gazebo Environment (a) \textit{ Wall Grid World} (b) \textit{Box Cylinder World} }
		\label{Gazebo_world}
\vspace{-7mm}
\end{figure}






\section{Conclusions and Future Work}
\vspace{-1mm}
In this paper, we made a fundamental contribution towards bringing the benefits of CCO-based safe trajectory planning to real-world settings commonly encountered in quadrotor navigation. Our CCO formulation works directly on the voxel grid representation of the world, wherein the collision constraints are not known in analytical form but rather in the form of a black-box query function. Existing approaches that assume known parametric nature in the underlying uncertainty and geometric collision avoidance models are not equipped to work under such minimalist assumptions. We outperformed state-of-the-art deterministic planners in success rate and smoothness metric. In the future, we aim to extend our work to handle state uncertainty in both static and dynamic environments.

\bibliography{references}
\footnotesize{
\bibliographystyle{IEEEtran}
}
\addtolength{\textheight}{-12cm}
\end{document}